\documentclass[11pt,a4paper]{article}
\usepackage{eacl2012}
\usepackage{times}
\usepackage{latexsym}
\usepackage{amsmath}
\usepackage{multirow}
\usepackage{graphicx}
\usepackage{url}

\title{ONTS: ``Optima'' News Translation System}

\author{Marco Turchi$^{*}$, Martin Atkinson$^{*}$, Alastair Wilcox$^{+}$, Brett Crawley, \\
\textbf{Stefano Bucci$^{+}$, Ralf Steinberger$^{*}$ and Erik Van der Goot$^{*}$}   \\
  European Commission - Joint Research Centre (JRC), IPSC - GlobeSec\\
Via Fermi 2749, 21020 Ispra (VA) - Italy\\
  {\tt $^{*}$[name].[surname]@jrc.ec.europa.eu}\\
   {\tt$^{+}$[name].[surname]@ext.jrc.ec.europa.eu }  \\
   {\tt brettcrawley@gmail.com }}
  
\date{}

\begin{document}
\maketitle
\begin{abstract}
We propose a real-time machine translation system that allows users to select a news category and to translate the related live news articles from Arabic, Czech, Danish, Farsi, French, German, Italian, Polish, Portuguese, Spanish and Turkish into English. The Moses-based system was optimised for the news domain and differs from other available systems in four ways: (1) News items are automatically categorised on the source side, before translation; (2) Named entity translation is optimised by recognising and extracting them on the source side and by re-inserting their translation in the target language, making use of a separate entity repository; (3) News titles are translated with a separate translation system which is optimised for the specific style of news titles; (4) The system was optimised for speed in order to cope with the large volume of daily news articles. 
%In this paper we present a complete translation system which translates large amounts of news items from several languages into English. This system is embedded in a family of news analysis applications which processes more than 100,000 news articles per day in more than fifty languages. Our service is based on the Phrase-Based Statistical Machine Translation toolkit Moses and is customised for the news translation task, taking advantage of the structure of the document and of additional information such as named entities.  We propose a real-time demo  which allows a user to select the news items by topic in one of these source languages, Arabic, Danish, Farsi, French, German, Italian, Polish, Spanish and Turkish, and to translate them into English. 
\end{abstract}

\section{Introduction}
Being able to read news from other countries and written in other languages allows readers to be better informed. It allows them to detect national news bias and thus improves transparency and democracy. Existing online translation systems such as \textit{Google Translate} and \textit{Bing Translator}\footnote{\url{http://translate.google.com/} and \url{http://www.microsofttranslator.com/}} are thus a great service, but the number of documents that can be submitted is restricted (Google will even entirely stop their service in 2012) and submitting documents means disclosing the users' interests and their (possibly sensitive) data to the service-providing company.

For these reasons, we have developed our in-house machine translation system ONTS. Its translation results will be publicly accessible as part of the Europe Media Monitor family of applications, \cite{Stei:09}, which gather and process about 100,000 news articles per day in about fifty languages. ONTS is based on the open source phrase-based statistical machine translation toolkit Moses~\cite{KoeHoa:07}, trained mostly on freely available parallel corpora and optimised for the news domain, as stated above. The main objective of developing our in-house system is thus not to improve translation quality over the existing services (this would be beyond our possibilities), but to offer our users a rough translation (a ``gist'') that allows them to get an idea of the main contents of the article and to determine whether the news item at hand is relevant for their field of interest or not.

A similar news-focused translation service is ``Found in Translation''~\cite{Tur:09}, which gathers articles in 23 languages and translates them into English. ``Found in Translation'' is also based on Moses, but it categorises the news after translation and the translation process is not optimised for the news domain.

\section{Europe Media Monitor}

Europe Media Monitor (EMM)\footnote{\url{http://emm.newsbrief.eu/overview.html}} gathers a daily average of 100,000 news articles in approximately 50 languages, from about 3,400 hand-selected web news sources, from a couple of hundred specialist and government websites, as well as from about twenty commercial news providers. It visits the news web sites up to every five minutes to search for the latest articles. When news sites offer RSS feeds, it makes use of these, otherwise it extracts the news text from the often complex HTML pages. All news items are converted to Unicode. They are processed in a pipeline structure, where each module adds additional information. Independently of how files are written, the system uses UTF-8-encoded RSS format.

Inside the pipeline, different algorithms are implemented to produce monolingual and multilingual clusters and to extract various types of information such as named entities, quotations,  categories and more. ONTS uses two modules of EMM: the named entity recognition and the categorization parts. %Several papers have been published on our news analysis system, but we do not mention them for reviewing purpose.

\subsection{Named Entity Recognition and Variant Matching.} \label{ner}
Named Entity Recognition (NER) is performed using manually constructed language-independent rules that make use of language-specific lists of trigger words such as titles (president), professions or occupations (tennis player, playboy), references to countries, regions, ethnic or religious groups (French, Bavarian, Berber, Muslim), age expressions (57-year-old), verbal phrases (deceased), modifiers (former) and more. These patterns can also occur in combination and patterns can be nested to capture more complex titles, \cite{Stei:07}. In order to be able to cover many different languages, no other dictionaries and no parsers or part-of-speech taggers are used. 

To identify which of the names newly found every day are new entities and which ones are merely variant spellings of entities already contained in the database, we apply a language-independent name similarity measure to decide which name variants should be automatically merged, for details see \cite{Pou:09}. This allows us to maintain a database containing over 1,15 million named entities and 200,000 variants. The major part of this resource can be downloaded from \url{http://langtech.jrc.it/JRC-Names.html}

\subsection{Category Classification across Languages.} \label{cat}
All news items are categorized into hundreds of categories. Category definitions are multilingual, created by humans and they include geographic regions such as each country of the world, organizations, themes such as natural disasters or security, and more specific classes such as earthquake, terrorism or tuberculosis,  

Articles fall into a given category if they satisfy the category definition, which consists of Boolean operators with optional vicinity operators and wild cards. Alternatively, cumulative positive or negative weights and a threshold can be used. Uppercase letters in the category definition only match uppercase words, while lowercase words in the definition match both uppercase and lowercase words. Many categories are defined with input from the users themselves. This method to categorize the articles is rather simple and user-friendly, and it lends itself to dealing with many languages, \cite{Stei:09}.

\section{News Translation System} \label{trans}
In this section, we describe our statistical machine translation (SMT) service based on the open-source toolkit Moses~\cite{KoeHoa:07} and its adaptation to translation of news items. 

\textbf{Which is the most suitable SMT system for our requirements?}  The main goal of our system is to help the user understand the content of an article. This means that a translated article is evaluated positively even if it is not perfect in the target language. Dealing with such a large number of source languages and articles per day, our system should take into account the translation speed, and try to avoid using language-dependent tools such as part-of-speech taggers.

Inside the Moses toolkit, three different statistical approaches have been implemented:  \textit{phrase based statistical machine translation} (PBSMT)~\cite{KoeOchMar:03}, \textit{hierarchical phrase based statistical machine translation}~\cite{Chi:07} and \textit{syntax-based statistical machine translation}~\cite{MarWan:06}. To identify the most suitable system for our requirements, we run a set of experiments training the three models with Europarl V4 German-English~\cite{Koe:05} and optimizing and testing on the News corpus~\cite{cal09}. For all of them, we use their default configurations and they are run under the same condition on the same machine to better evaluate translation time. For the syntax model we use linguistic information only on the target side. According to our experiments, in terms of performance the hierarchical model performs better than PBSMT and syntax (18.31, 18.09, 17.62 Bleu points), but in terms of translation speed PBSMT is better than hierarchical and syntax (1.02, 4.5, 49 second per sentence). Although, the hierarchical model has the best Bleu score, we prefer to use the PBSMT system in our translation service, because it is four times faster.

%Translation time, second per sentence, and quality, BLEU score, are shown in Table. 

%\begin{table}[h]
%\begin{center}
%\begin{tabular}{|l|cc|}
%\hline \bf System & \bf Trans. Time & \bf Bleu Score \\ \hline
%PBSMT & \textbf{1.02}  & 18.09 \\
%HIERO & 4.5 pt & \textbf{18.31} \\
%Syntax & 49  & 17.62\\
%\hline
%\end{tabular}
%\end{center}
%\caption{\label{font-table} Font guide. }
%\end{table}

\textbf{Which training data can we use?} It is known in statistical machine translation that more training data implies better translation. Although, the number of parallel corpora has been is growing in the last years, the amounts of training data vary from language pair to language pair. To train our models we use the freely available corpora (when  possible): Europarl~\cite{Koe:05}, JRC-Acquis~\cite{Stei:06}, DGT-TM\footnote{\url{http://langtech.jrc.it/DGT-TM.html}}, Opus~\cite{Tie:09}, SE-Times~\cite{TyeAlp:10}, Tehran English-Persian Parallel Corpus~\cite{Pil:11}, News Corpus~\cite{cal09}, UN Corpus~\cite{Raf:09}, CzEng0.9~\cite{Bo:2009},  English-Persian parallel corpus distributed by ELRA\footnote{\url{http://catalog.elra.info/}} and two Arabic-English datasets distributed by LDC\footnote{\url{http://www.ldc.upenn.edu/}}. This results in some language pairs with a large coverage, (more than 4 million sentences), and other with a very small coverage, (less than 1 million). The language models are trained using 12 model sentences for the content model and 4.7 million  for the title model. Both sets are extracted from English news.

For less resourced languages such as Farsi and Turkish, we tried to extend the available corpora. For Farsi, we applied the methodology proposed by~\cite{Lam:11}, where we used a large language model and an English-Farsi SMT model to produce new sentence pairs. For Turkish we added the Movie Subtitles corpus~\cite{Tie:09}, which allowed the SMT system to increase its translation capability, but included several slang words and spoken phrases.

\textbf{How to deal with Named Entities in translation?} News articles are related to the most important events.  These names need to be efficiently translated to correctly understand the content of an article. From an SMT point of view, two main issues are related to Named Entity translation: (1) such a name is not in the training data or (2) part of the name is a common word in the target language and it is wrongly translated, e.g. the French name ``Bruno Le Maire'' which risks to be translated into English as ``Bruno Mayor''. 
 To mitigate both the effects we use our multilingual named entity database. In the source language, each news item is analysed to identify possible entities; if an entity is recognised, its correct translation into English is retrieved from the database, and suggested to the SMT system enriching the source sentence using the xml markup option \footnote{\url{http://www.statmt.org/moses/?n=Moses.AdvancedFeatures#ntoc4}} 
 in Moses. This approach allows us to complement the training data increasing the translation capability of our system.
 
\textbf{How to deal with different language styles in the news?} News title writing style  contains more gerund verbs, no or few linking verbs, prepositions and adverbs than normal sentences, while content sentences include more preposition, adverbs and different verbal tenses. Starting from this assumption, we investigated if this phenomenon can affect the translation performance of our system. 

We trained two SMT systems, $SMT_{content}$ and $SMT_{title}$, using the Europarl V4 German-English data as training corpus, and two different development sets: one made of content sentences, News Commentaries~\cite{cal09}, and the other made of news titles in the source language which were translated into English using a commercial translation system. With the same strategy we generated also a Title test set. The $SMT_{title}$ used a language model created using only English news titles.  The News and Title test sets were translated by both the systems.  Although the performance obtained translating the News and Title corpora are not comparable, we were interested in analysing how the same test set is translated by the two systems. We noticed that translating a test set with a system that was optimized with the same type of data resulted in almost 2 Blue score improvements:  Title-TestSet: 0.3706 ($SMT_{title}$), 	0.3511 ($SMT_{content}$); News-TestSet: 0.1768 ($SMT_{title}$), 0.1945 ($SMT_{content}$).  This behaviour was present also in different language pairs. According to these results we decided to use two different translation systems for each language pair, one optimized using title data and the other using normal content sentences. Even though this implementation choice requires more computational power to run in memory two Moses servers, it allows us to mitigate the workload of each single instance reducing translation time of each single article and to improve translation quality.

\subsection{Translation Quality}
To evaluate the translation performance of ONTS, we run a set of experiments where we translate a test set for each language pair  using our system and Google Translate. Lack of human translated parallel titles obliges us to test only the content based model. For German, Spanish and Czech we use the news test sets proposed in \cite{cal10}, for French and Italian the news test sets presented in \cite{cal08}, for Arabic, Farsi and Turkish, sets of 2,000 news sentences extracted from the Arabic-English and English-Persian datasets and the SE-Times corpus.  For the other languages we use 2,000 sentences which are not news but a mixture of JRC-Acquis, Europarl and DGT-TM data. It is not guarantee that our test sets are not part of the training data of Google Translate.

Each test set is translated by Google Translate - Translator Toolkit, and by our system. Bleu score is used to evaluate the performance of both systems. Results, see Table \ref{results}, show that Google Translate produces better translation for those languages for which large amounts of data are available such as French, German, Italian and Spanish. Surprisingly, for Danish, Portuguese and Polish, ONTS has better performance, this depends on the choice of the test sets which are not made of news data but of data that is fairly homogeneous in terms of style and genre with the training sets.  

The impact of the named entity module is evident for Arabic and Farsi, where each English suggested entity results in  a larger coverage of the source language and better translations.   For highly inflected and agglutinative languages such as Turkish, the output proposed by ONTS is poor. We are working on gathering more training data coming from the news domain and on the possibility of applying a linguistic pre-processing of the documents.

%
%\begin{table}[h]
%\begin{center}
%\begin{tabular}{|l|cc|cc|}
%\hline 
%%\bf Source  & \bf ONTS &  \bf Google \\ 
%% \bf  Language & \bf  & \bf  \\\hline
%\bf Source L. & \multicolumn{2}{|c|}{\bf ONTS} & \multicolumn{2}{|c|}{\bf Google T.} \\ \hline
% & B & B-C & B & B-C \\
%Arabic & 0.318 & -  & 0.255 & - \\
%Czech & 0.218 & 0.149 & 0.226  & 0.18 \\
%Danish & 0.324  & 0.234 &  0.296  & 0.249  \\
%Farsi & 0.245  & 0.174 & 0.197 & 0.16\\
%French & 0.26 & 0.193 & 0.286& 0.234\\
%German & 0.205 & 0.135 & 0.25 & 0.204\\
%Italian & 0.234 & 0.176 & 0.31& 0.261\\
%Polish & 0.568 & 0.391 & 0.511 & 0.443\\
%Portuguese & 0.579 & 0.424 & 0.424 & 0.396 \\
%Spanish & 0.283 & 0.205 & 0.334 & 0.273\\
%Turkish & 0.238 & 0.132 & 0.395 & 0.344\\
%\hline
%\end{tabular}
%\end{center}
%\caption{\label{font-table} Automatic evaluation. B: Bleu score on lowercased and tokenized data. B-C: Bleu score on cased and detokenized data. Arabic test set available only tokenized and lowercased.}
%\end{table}

\begin{table}[h]
\begin{center}
\begin{tabular}{|l|c|c|}
\hline 
%\bf Source  & \bf ONTS &  \bf Google \\ 
% \bf  Language & \bf  & \bf  \\\hline
\bf Source L. &\bf ONTS &\bf Google T. \\ \hline
% & Bleu & Bleu \\
Arabic & 0.318  & 0.255 \\
Czech & 0.218  & 0.226   \\
Danish & 0.324   &  0.296   \\
Farsi & 0.245   & 0.197 \\
French & 0.26  & 0.286\\
German & 0.205  & 0.25 \\
Italian & 0.236  & 0.31\\
Polish & 0.568  & 0.511 \\
Portuguese & 0.579  & 0.424  \\
Spanish & 0.283 & 0.334 \\
Turkish & 0.238  & 0.395 \\
\hline
\end{tabular}
\end{center}
\caption{\label{results} Automatic evaluation.}
\end{table}

%[Marco]
%\begin{itemize}
%\item Choice of the SMT system
%\item Named Entity Adaptation
%\item News Customisation
%\item Data and Low Resourced Languages
%\end{itemize}

\section{Technical Implementation} \label{tech}
The translation service is made of two components: the connection module and the Moses server. The connection module is a servlet implemented in Java. It receives the RSS files, isolates each single news article, identifies each source language  and pre-processes it. Each news item is split into sentences, each sentence is tokenized, lowercased, passed through a statistical compound word splitter, \cite{KoeKni:03}, and the named entity annotator module. For language modelling we use the KenLM implementation, \cite{Hea:11}.

According to the language, the correct Moses servers, title and content, are fed in a multi-thread manner. We use the multi-thread version of Moses~\cite{Had:10}. When all the sentences of each article are translated, the inverse process is run: they are detokenized, recased, and untranslated/unknown words are listed. The translated title and content of each article are uploaded into the RSS file and it is passed to the next modules. 

The full system including the translation modules is running in a 2xQuad-Core with Intel Hyper-threading Technology processors with 48GB of memory. It is our intention to locate the Moses servers on different machines. This is possible thanks to the high modularity and customization of the connection module. At the moment, the translation models are available for the following source languages: Arabic, Czech, Danish, Farsi, French, German, Italian, Polish, Portuguese, Spanish and Turkish.

\subsection{Demo}
\begin{figure}[t]
\centering
\includegraphics[width=0.49\textwidth]{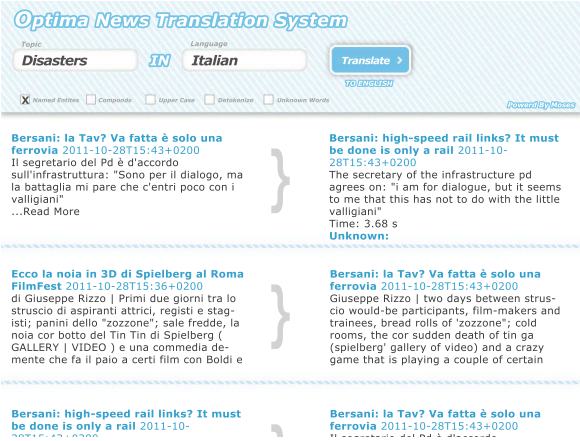}%[scale=0.35]{./figure/moses_layout_01.jpg}

\caption{Demo Web site.}
\label{Fig::Demo}
\end{figure}

Our translation service is currently presented on a demo web site, see Figure \ref{Fig::Demo}, which is available at \url{http://optima.jrc.it/Translate/}. News articles can be retrieved selecting one of the topics and the language. All the topics are assigned to each article using the methodology described in \ref{cat}. These articles are shown in the left column of the interface.  %A user needs to select the articles that have to be translated clicking on each of them. 
When the button ``Translate'' is pressed, the translation process starts and the translated articles appear in the right column of the page. 

The translation system can be customized from the interface enabling or disabling the named entity, compound, recaser, detokenizer and unknown word modules. Each translated article is enriched showing the translation time in milliseconds per character and, if enabled, the list of unknown words. The interface is linked to the connection module and data is transferred using RSS structure.  %The preliminary demo of ONTS at \url{} will be replaced by a final version before EACL 2012. %Our demo will be accessible from outside our network for the conference.

\section{Discussion} \label{concl}
In this paper we present the Optima News Translation System and how it is connected to Europe Media Monitor application. Different strategies are applied to increase the translation performance taking advantage of the document structure and other resources available in our research group. We believe that the experiments described in this work can result very useful for the development of other similar systems. Translations produced by our system will soon be available as part of the main EMM applications. 

 The performance of our system is encouraging, but not as good as the performance of web services such as Google Translate, mostly because we use less training data and we have reduced computational power. On the other hand, our in-house system can be fed with a large number of articles per day and sensitive data without including third parties in the translation process.  Performance and translation time vary  according to the number and complexity of sentences and language pairs. 

%On average, it takes few milliseconds per character to translate a full article. This varies according to the number and complexity of sentences and language pairs. 

The domain of news articles dynamically changes according to the main events in the world, while existing parallel data is static and usually associated to governmental domains. It is our intention to investigate how to adapt our translation system  updating the language model with the English articles of the day. 

\section*{Acknowledgments}
The authors thank the JRC's OPTIMA team for its support during the development of ONTS.

\end{document}